# Pessimistic asynchronous sampling in high-cost Bayesian optimization

Amanda A. Volk[1]; Kristofer G. Reyes[2], Jeffrey G. Ethier[3], Luke A. Baldwin[3]

[1] *National Research Council Research Associate, Air Force Research Laboratory, Wright-Patterson Air Force Base, OH, 45433, United States*

[2] *Department of Materials Design and Innovation, University at Buffalo, Buffalo, NY 14260*

[3] *Materials and Manufacturing Directorate, Air Force Research Laboratory, Wright-Patterson Air Force Base, OH 45433, United States*

**Abstract**

Asynchronous Bayesian optimization is a recently implemented technique that allows for parallel operation of experimental systems and disjointed workflows. Contrasting with serial Bayesian optimization which individually selects experiments one at a time after conducting a measurement for each experiment, asynchronous policies sequentially assign multiple experiments before measurements can be taken and evaluate new measurements continuously as they are made available. This technique allows for faster data generation and therefore faster optimization of an experimental space. This work extends the capabilities of asynchronous optimization methods beyond prior studies by evaluating four additional policies that incorporate pessimistic predictions in the training data set. Combined with a conventional policy that uses model predictions, the five total policies were evaluated in a simulated environment and benchmarked with serial sampling. Under some conditions and parameter space dimensionalities, the pessimistic prediction asynchronous policy reached optimum experimental conditions in significantly fewer experiments than equivalent serial policies and proved to be less susceptible to convergence onto local optima at higher dimensions. Without accounting for the faster sampling rate, the pessimistic asynchronous algorithm presented in this work could result in more efficient algorithm driven optimization of high-cost experimental spaces. Accounting for sampling rate, the presented asynchronous algorithm could allow for faster optimization in experimental spaces where multiple experiments can be run before results are collected.

## Introduction

Asynchronous Bayesian optimization algorithms enable greater flexibility in algorithm assisted experimental workflows. In most traditional scientific experimental procedures, experiments are conducted over distinct process steps. For example, in reaction chemistry research, a reaction process is often conducted in one experimental apparatus and the synthesized material is characterized with a separate analysis tool. In many scenarios, such as workflows leveraging transmission/scanning electron microscopy, x-ray diffraction, or even high-field nuclear magnetic resonance spectroscopy, the characterization process can take more time than the experiment. For any single experiment, some portion of the equipment is available for use before the completion of the procedure, which means that an algorithm would have the opportunity to select additional tests to run before it has data from the prior experiment. This issue is further compounded in experimental environments that rely on a human within the experiment conduction and selection loop. Although this limitation is less prevalent in fully closed-loop experimental contexts, many platform designs and experimental systems would benefit from running multiple experiments or experimental steps simultaneously.[5–9]

One strategy for resolving an incomplete utilization of resources is batch, also referred to as parallel sampling. In batch sampling, a set of experiments are defined and conducted with complete utilization of parallelized experimental resources during each stage of an experimental process, then the measurements from that set of experiments are simultaneously returned to the algorithm for selection of the next set of experiments.[10] This approach is suitable for select experimental environments, such as combinatorial screening platforms or high time cost measurements. However, batch sampling poses several intuitive issues in sampling efficiency. First, while equipment utilization is improved, there is typically still equipment down time when alternating between the different stages of the experiments. Second, batch sampling often does not maximize data availability in algorithm decision making. Unless the experimental system is inherently structured for batch sampling, there is typically a missed opportunity to complete an experiment and measurement that informs the experiment selection algorithm before conducting all the experiments in the set. Finally, batch methods are not suitable for experimental systems with time dependent outcomes. For example, if an experiment were to produce a material that degrades over time, batch methods would not result in a uniform time step between experiment and measurement, resulting in imprecise data generation.

In response to the constraints of batch sampling strategies, asynchronous sampling methods have recently been implemented in high-cost experimental environments,[11] specifically in delocalized experimentation networks.[12,13] Shown in Figure 1, asynchronous sampling methods implement similar strategies to batch sampling by selecting multiple experiments without completing measurements. However, in asynchronous designs, experiments are continuously measured and added to the data set while other experimental steps are being conducted. In an asynchronous Bayesian optimization design, there is a moving window buffer that contains placeholder data for the currently running experiments. This buffer set is appended to the real value data set for model training. When an experiment measurement is complete, the real data replaces the placeholder data. Then, a new experiment is selected, and the placeholder data is added to the buffer. Several strategies have been implemented to generate placeholder values in asynchronous Bayesian optimization, including local penalty strategies[14–17] and realistic constant liar predictions[11] among others.[18,19] In prior studies, asynchronous sampling resulted in faster data generation rates and therefore faster approach to optimal experimental conditions.

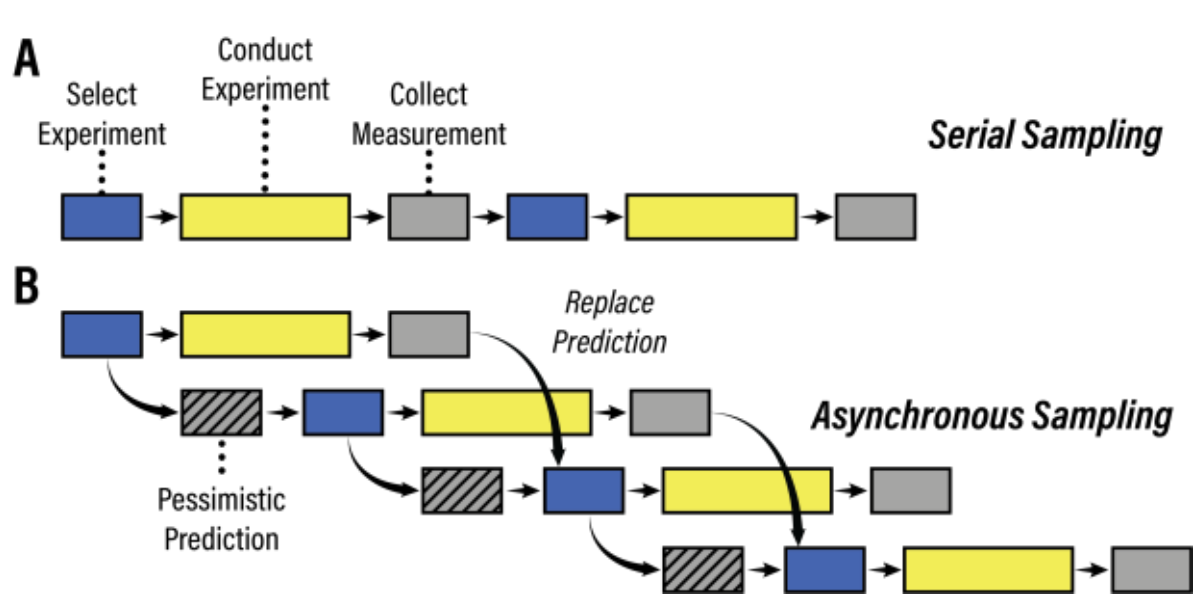

**Figure 1.** Illustration of asynchronous policy selection process versus serial selection.

In this work we present five alternative asynchronous sampling value prediction policies: (1) pessimistic constant liar prediction, (2) random prediction, (3) descending pessimism constant liar prediction, (4) ascending pessimism constant liar prediction, and (5) lower confidence bounds prediction. In each of these policies, we explore methods for selecting the values used in the placeholder prediction buffer. We benchmark these four policies with serial sampling and realistic prediction asynchronous sampling on a selection of representative surrogate ground truth functions. The simulated optimization campaigns on surrogate functions showed that with an upper confidence bounds decision policy and a Gaussian Process regressor, the realistic prediction policy and all three alternative policies outperformed serial sampling considerably when accounting for the improved sampling rate. Furthermore, we found that all five alternative policies consistently performed competitively with serial sampling and in some cases significantly outperformed serial sampling when accounting for the number of experiments conducted. The pessimistic constant liar prediction was also found to outperform serial sampling on discrete input spaces of real-world data modeled with a random forest regressor. The three proposed strategies not only generate data at a faster rate than serial sampling, but they also select experiments equally or more efficiently. Implementation of the proposed algorithm has notable implications in asynchronous experiment conduction loops for high-cost experiments, and it could potentially improve the sampling efficiency of serial closed-loop systems.

## Methods

*Decision Policy*

The asynchronous Bayesian optimization simulations were conducted using a modified BOBnd framework.[20] In all simulations, the surrogate model was a scikit-learn[21] Gaussian Process regressor with a radial basis function kernel and a limited-memory BFGS optimizer, and the acquisition function was an upper confidence bounds policy as defined below:

$$\boldsymbol{X}_{Next} = argmin\big(y'(\boldsymbol{X}) + \lambda\sigma(\boldsymbol{X})\big), \qquad \lambda = \frac{5}{\sqrt{2}}$$

$$\boldsymbol{X} = [x_1, x_2, \dots x_D]$$

where $\boldsymbol{X}_{Next}$ is the next set of experimental conditions, $y'$ and $\sigma$ are the model response mean prediction and standard deviation respectively as a function of the experimental conditions vector $\boldsymbol{X}$, $x_i$ is the condition input value for dimension $i$, $D$ is the total number of dimensions in the input space, and $\lambda$ is the exploration constant. The value of $\lambda$ was selected using a preliminary screening of the pessimistic and realistic asynchronous sampling policies, shown in Supporting Information Figure S.1, and selected for generating the highest performing algorithm on five-dimensional *TriPeak*. Note that the distinctions between serial and pessimistic asynchronous policies discussed in this work are less prominent at higher $\lambda$ values. However, using pessimistic asynchronous sampling, performance is less dependent on $\lambda$ values in the UCB policy. Simulations were also carried out with dynamic exploration constants[22] on additional surrogate functions from prior literature, shown in Supporting Information Figure S.2. Even with this dimension-scaled exploration constant, pessimistic policies matched or outperformed realistic prediction policies at higher buffer length values.

*Asynchronous Predictions*

The asynchronous sampling policy generates a vector of predicted values, referred to as the buffer array ($B$), for all incomplete experiments. For the simulations conducted in this work, the length of the buffer array ($N_{Buff}$) varied between one to ten predicted samples, corresponding to one to ten simultaneously running experiments. Each of the six buffer policies – realistic, pessimistic, random, ascending pessimism, descending pessimism, and lower confidence bounds – fill in the buffer arrays $B_{Real}$, $B_{Pess}$, $B_{Rand}$, $B_{AscPess}$, $B_{DesPess}$, and $B_{LCB}$ respectively – with the following equations:

$$B_{Real} = \langle y'(X_{C+1}), y'(X_{C+2}), \dots y'\left(X_{C+NBuff}\right)\rangle$$

$$B_{Pess} = \langle 0, 0, \dots 0\rangle$$

$$B_{Rand} = \langle u_1, u_2, \dots u_{NBuff}\rangle \; where \; u_i \sim \mathcal{U}(0, 1)$$

$$B_{AscPess} = \left\langle \begin{matrix} \frac{N_{Buff} - 1}{N_{Buff}} y'(X_{C+1}), \\ \frac{N_{Buff} - 2}{N_{Buff}} y'(X_{C+2}), \\ \dots \frac{N_{Buff} - N_{Buff}}{N_{Buff}} y'\left(X_{C+NBuff}\right) \end{matrix} \right\rangle$$

$$B_{DecPess} = \left\langle \begin{matrix} \frac{0}{N_{Buff}} y'(X_{C+1}), \\ \frac{1}{N_{Buff}} y'(X_{C+2}), \\ \dots \frac{N_{Buff}}{N_{Buff}} y'\left(X_{C+NBuff}\right) \end{matrix} \right\rangle$$

$$B_{LCB} = \begin{matrix} \langle argmin\left(y'(X_{C+1}) - \lambda\sigma(X_{C+1})\right), \\ argmin\left(y'(X_{C+2}) - \lambda\sigma(X_{C+2})\right), \\ \dots \; argmin\left(y'\left(X_{C+NBuff}\right) - \lambda\sigma\left(X_{C+NBuff}\right)\right)\rangle \end{matrix}$$

where $X_{C+j}$ is the input vector for the buffer position $(j)$ after the most recent completed experiment $(C)$ and $\mathcal{U}(0, 1)$ is a uniform random distributed that is bounded from 0 to 1. A pessimistic value is defined as the lower bound of the expected response range, which in the case of the *TriPeak* surrogate is zero. The pessimistic assumption, also referred to as censorship in prior works,[23] has been leveraged in multi-worker contexts where delay distributions are randomly sampled to dynamically determine the buffer lengths, but it has not been evaluated under uniform delay asynchronous sampling. The uniform random distribution is resampled for every value in the buffer each time the model is trained.

*Surrogate Ground Truth*

The surrogate ground truth function, $f(X)$, referred to as *TriPeak*, is an N-dimensional, triple Gaussian peak integrand function adapted from the BOBnd library,[20] Surjanovic and Bingham,[24] and Genz,[25] and is defined with the equation below:

$$f(X) = c\left(b_1 \exp\left(\sum_{i=1}^{D} a_1^2(x_i - \mu_1)^2\right) + b_2 \exp\left(\sum_{i=1}^{D} a_2^2(x_i - \mu_2)^2\right) + b_3 \exp\left(\sum_{i=1}^{D} a_3^2(x_i - \mu_3)^2\right)\right)$$

$$c = \frac{1}{D(b_1 + b_2 + b_3)}$$
$$a_1 = 4, \quad \mu_1 = 0.2, \quad b_1 = 0.3$$
$$a_2 = 1.5, \quad \mu_2 = 0.5, \quad b_2 = 0.2$$
$$a_3 = 4, \quad \mu_3 = 0.8, \quad b_3 = 0.6$$

where $c$ is the normalization scalar, $a_j$ is the peak width modifier, $\mu_j$ is the peak location, and $b_j$ is the peak height modifier for a single dimension of peak $j$. In the noisy simulations, the ground truth value was sampled with random noise added by sampling from a normal probability distribution with a mean of zero and standard deviation specified by a noise value of 0.01, 0.02, and 0.05 for 1%, 2% and 5% noise respectively.

Simulations using additional surrogate functions are reported in the supporting information along with randomized sampling control groups, shown in Supporting Information Figure S.3.

**Results**

The five asynchronous sampling policies were studied with a five-dimensional *TriPeak* surrogate ground truth function, with the results shown in Figure 2 and S.4. All five policies demonstrated some viability in accelerating optimization rates through parallel experimentation. However, the realistic prediction policy exhibited significant losses in sampling efficiency when increasing the buffer length to four samples or higher. Additionally, no asynchronous buffer length with a realistic policy outperformed serial sampling when evaluated as a function of the number of experiments. Among the four tested policies, a pure pessimistic policy had the highest and most consistent performance across all campaign replicates and buffer lengths, and the descending pessimism performed similarly. While the ascending pessimism policy performed more favorably than the realistic policy across all buffer lengths, the highest buffer length showed some indication of a less consistent or slower convergence onto the optimum. Similarly, random predictions showed comparable performance to pessimistic but the optimization efficiency over serial is lost at 9 buffers.

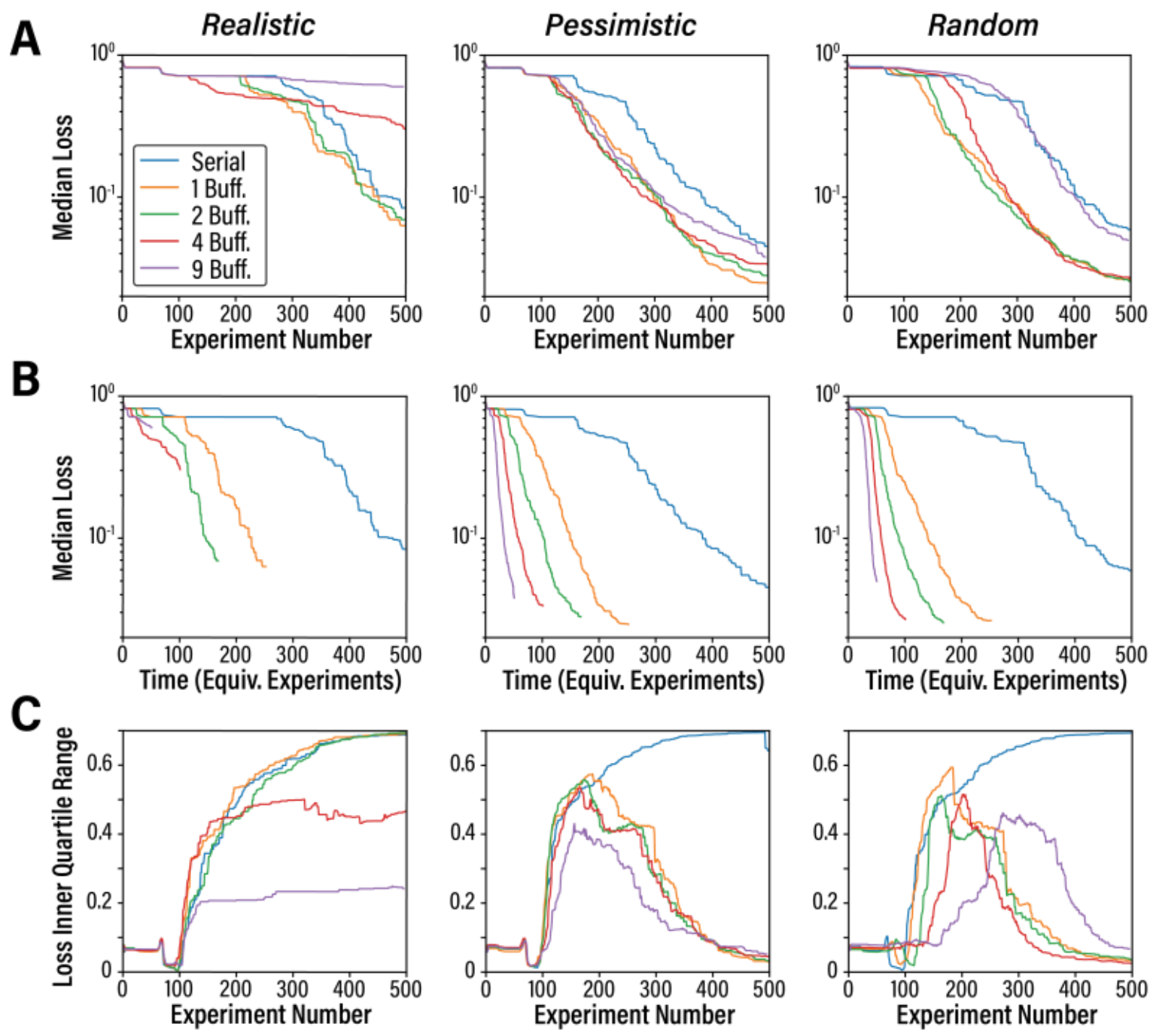

**Figure 2.** Simulation results of four asynchronous decision policies utilizing different prediction strategies on five-dimensional *TriPeak*. The median loss across all 200 randomized simulated campaigns as a function of (A) the number of experiments and (B) the effective optimization time relative to a single experiment and (C) the inner quartile range of the loss as a function of experiment number across the four decision policies, (first column) realistic, (second column) pessimistic, and (third column) random. The serial replicates were repeated for each of the four policies.

Both the pessimistic and descending pessimism policies outperformed serial sampling for all tested buffer lengths. After 500 experiments, all buffer lengths in the descending pessimism policy reached approximately 60% of the final loss achieved in serial sampling. For the pessimistic policy the 1, 2, 4, and 9 buffer campaigns reached equivalent performance to the serial campaign with 500 experiments in approximately 380, 380, 410, and 480 experiments respectively. After 500 experiments, all asynchronous policies leveraging pessimism showed a significant reduction in the inner quartile range after five hundred experiments, with the pessimistic policy reaching over an order of magnitude lower inner quartile range than the serial trial. The realistic prediction and serial policies continued to increase or plateau after 500 experiments. These findings suggest that the presence of pessimism in asynchronous policies produces more effective and consistent optimization and increasing the range of those pessimistic predictions can further increase the consistency with which campaigns reach optimal conditions. Additionally, the greatest algorithm improvement is observed after the inclusion of a single pessimistic prediction, i.e. one buffer length for the pessimistic, ascending pessimism, and descending pessimism policies. Significant improvements are observed with modest additions of pessimistic predictions, while realistic predictions either have no impact or decrease algorithm performance. Shown in Supporting Information Figure S.5, the constant buffer length implementation outperformed a simulated scenario where buffer lengths are changing dynamically between zero and the specified buffer throughout the optimization. Additionally, the prediction of pessimism near predicted optima is shown to be important as the introduction of randomized pessimism on a realistic buffer policy performed significantly worse than the pessimistic policy.

Similar trends can be observed when using different methods for selecting the exploration constant in the upper confidence bounds policy, shown in Supporting Information Figures S.1 and S.2. Pessimism generates the most significant improvement when lower exploration constants are used, and it influences efficiency less significantly when higher exploration constants are used. Additionally, when a more robust and dynamic exploration term is used, the improvement of pessimism over realistic policies is reduced further, indicating that higher exploration rates reduce the effectiveness of the pessimistic policy. Finally, applying pessimism with dynamic exploration terms on high complexity surrogates and very low complexity surrogates generates negligible improvement for most conditions. Across all lambda strategies and surrogate functions, with the one exception of the very simple surrogate function *Trid*, the pessimistic policy performed either equivalently or better than serial and realistic policies as a function of experiment number. As shown in Supporting Information Figure S.3, the dynamic exploration term policies performed worse than random sampling for the high complexity and very low complexity surrogates. In all scenarios where the decision policies outperformed random sampling, the pessimistic policy generated a substantial improvement in performance at the highest tested buffer length. When comparing all buffer lengths by experimental time rather than number, the pessimistic policy significantly outperforms the realistic asynchronous sampling.

The lower confidence bounds policy performed equivalently to the pessimistic policy when evaluated over one and two buffer lengths; however, the policy appears to converge prematurely and perform worse than the serial policy at high buffer lengths. Small buffer lower confidence bounds policies likely behave similarly to pessimistic policies in that the uncertainty near local optima is high enough to provide a sufficiently pessimistic hallucination. The failure at higher buffer lengths could be attributed to excessively confident models near local optima where clusters of buffer experiments are selected. In this latter case, the policy likely behaves more similarly to the realistic policy and provides insufficient pessimism to encourage exploration.

This pessimistic asynchronous method also demonstrates higher performance relative to serial sampling at higher dimensionalities. Shown in Figure 3, the serial policy outperforms all asynchronous pessimistic policies as a function of experiment number for two, three, and four-dimensional surrogate spaces. However, the asynchronous policies considerably outperformed the serial method for five and six dimensional spaces as shown in the fourth and fifth columns of Figure 3. In the six-dimensional space, the serial sampling policy improved very little after 2,000 experiments, while the pessimistic asynchronous policies approached optimal conditions. The asynchronous method provided a performance advantage for all five studied dimensions when considered as a function of experimentation time.

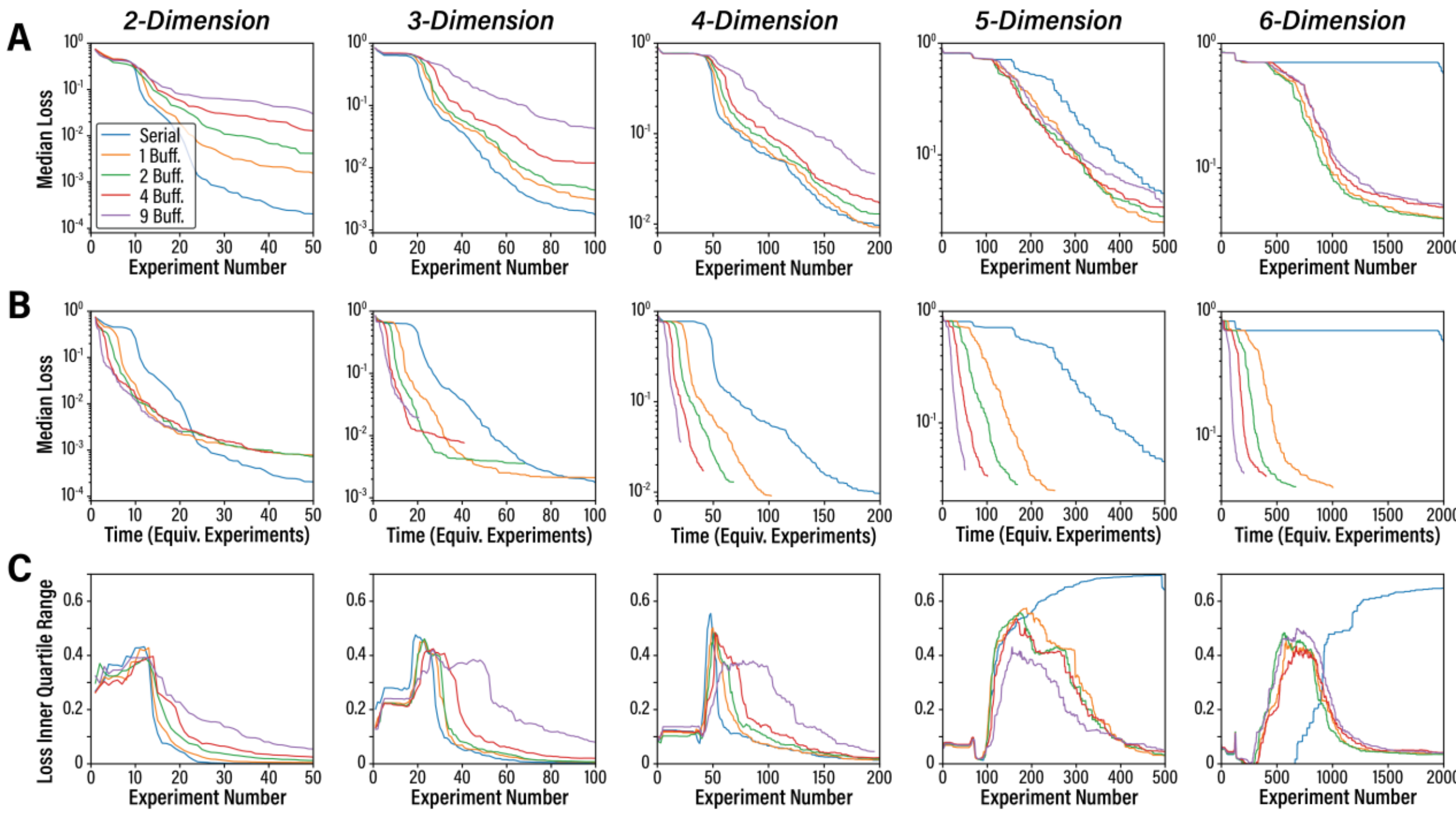

**Figure 3.** Simulation results of pessimistic decision policies on *TriPeak* at different dimensionalities and buffer lengths. The median loss across all randomized simulated campaigns as a function of (A) the number of experiments and (B) the effective optimization time relative to a single experiment and (C) the inner quartile range of the loss as a function of experiment number across (columns) two, three, four, five, and six-dimensional surrogate spaces. Each dimensional plot is the result of 200 replicates.

One potential explanation for the efficacy of pessimism assisted asynchronous sampling strategies is that the pessimistic predictions reduce the occurrence of premature convergence in upper confidence bounds policies. By forcing a pessimistic prediction on what the current model indicates is the optimal condition prevents resampling at that point, and in cases where replicates already exist outside the buffer, it increases model uncertainty at that point to enable improved exploration within the peak. This advantage becomes more dominant when the number of local minima (i.e., the dimensionality of the *TriPeak* function) increases.

The integration of the pessimistic prediction within the model training data set contrasts with prior pessimistic prediction methods on constant buffer length systems which implement a penalty region over a defined area around the prior data point. It is possible that these penalty region methods could suffer from the curse of dimensionality as the volume covered by the defined penalty areas represents a smaller fraction of the overall parameter space.[26]

A final study was conducted by introducing noise on the five-dimensional *TriPeak* surrogate ground truth function using the Pessimistic buffer policy across two to six dimensions. Shown in Figure 4, increasing the noise of the surrogate system resulted in less efficient optimization algorithms in most cases, but the serial policy at higher dimensions gained a performance advantage likely due to the normalization effect of noisy sampling. Like the no noise simulations, the serial policy outperformed the asynchronous policies for all noise levels at lower dimensionality, and the magnitude of the sampling penalty increased as the buffer size increased. Additionally, the introduction of noise negates any advantage with respect to experiment number attained by the buffer policies at five and six dimensions; however, the asynchronous policies substantially overlap with the results of the serial policy with respect to experiment number at these higher dimensions. This result further supports the notion that large buffers in pessimistic asynchronous sampling algorithms can provide faster optimizations with no or negligible impact on experimental efficiency.

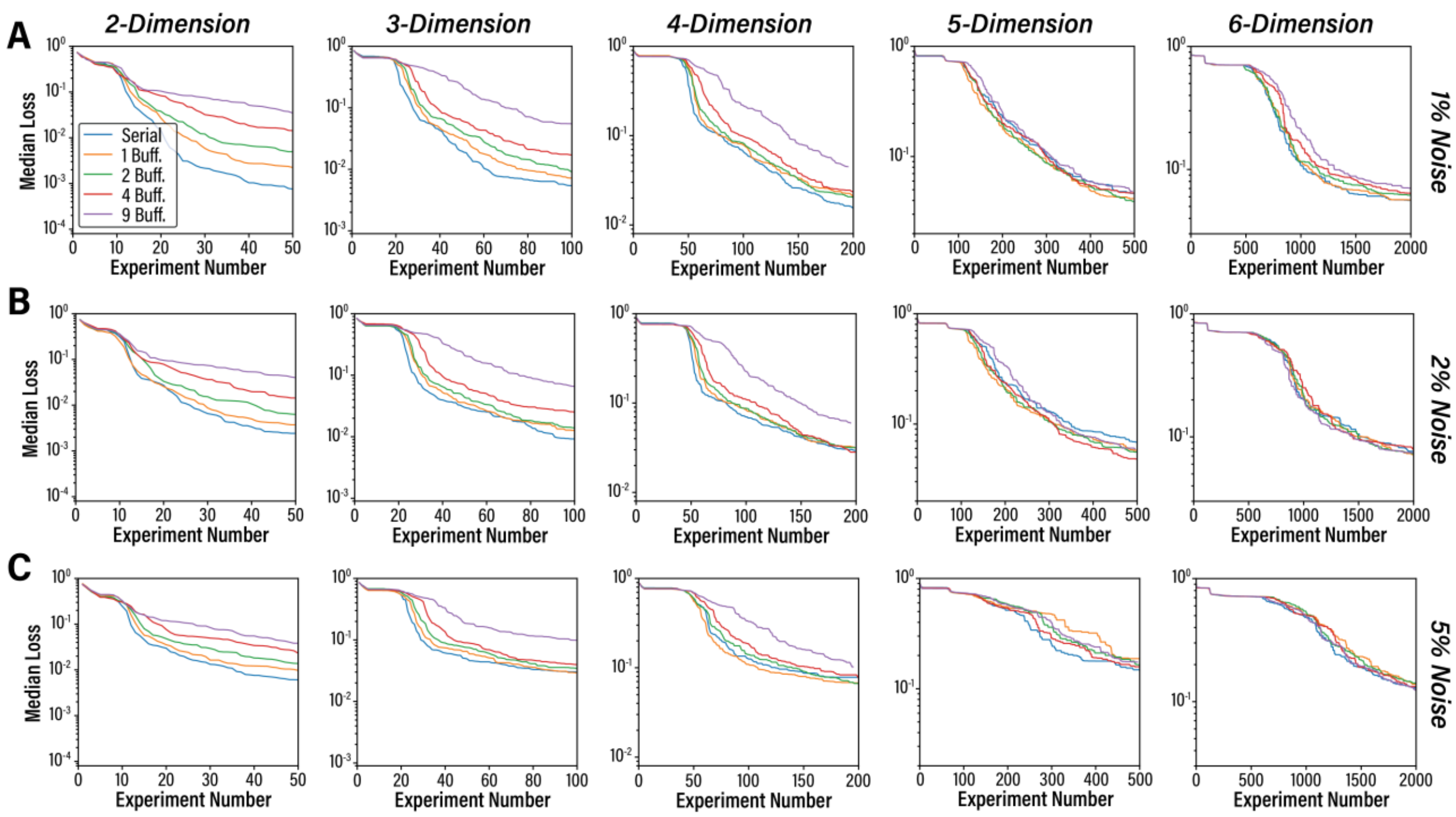

**Figure 4.** Simulation results of pessimistic decision policies on *TriPeak* at different dimensionalities and noise levels. The median loss across all randomized simulated campaigns as a function of the number of experiments across (columns) two, three, four, five, and six-dimensional surrogate spaces with (A) 1%, (B) 2%, and (C) 5% surrogate noise. Sampling noise is applied by randomly sampling from a normal distribution with a standard deviation equal to the specified noise value and adding the noise sample to the surrogate ground truth output. Each dimensional plot is the result of 200 replicates. The loss is calculated from the noiseless ground truth and does not reflect the values sampled from the surrogate during each trials campaign.

By implementing pessimistic predictions through model integration, the asynchronous sampling policies presented here could more effectively navigate higher dimension parameter spaces through more efficient and comprehensive integration of pessimism. In the complex experimental spaces relevant to algorithm driven experimentation, asynchronous policies provide a notable advantage to serial algorithms when parallel operation is viable. Furthermore, pessimistic asynchronous policies may provide an additional advantage over realistic based hallucinations.

### *Asynchronous Policy on Cross-Coupling Reaction Data*

Finally, the pessimistic asynchronous sampling algorithm was evaluated using a real-world data set. In Ahneman et al. [27] a comprehensive database with measured yield of a C-N cross-coupling reaction was generated for every combination of 15 aryl halides, 23 additives, 4 catalysts, and 3 bases. In total this database provides a measured yield for approximately 4,000 reactions. Additionally, the work converted the categorical space to a discrete numerical space by providing vector arrays based on molecular descriptors of each of the reactants. As a result, this database contains 120 total parameters that go into the yield measurements. In a simulated campaign, this data set was used as a surrogate function for algorithmically optimizing the reaction yield – shown in Figure 5A.

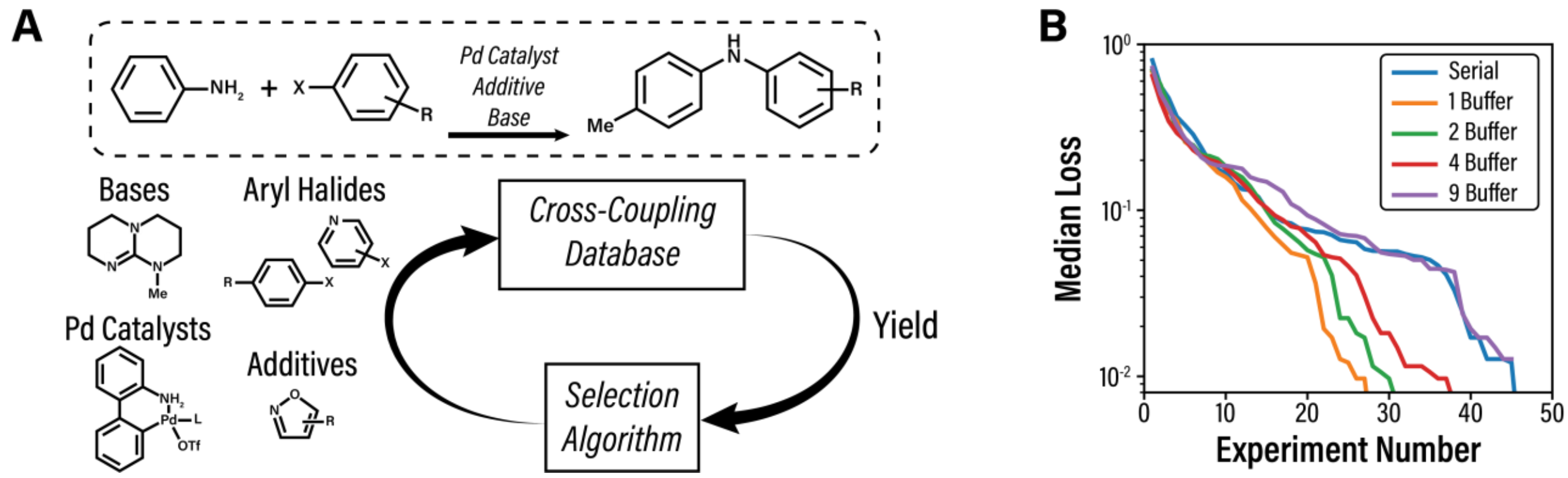

**Figure 5.** Asynchronous optimization on real C-N cross-coupling data. (A) Illustration of simulation design setup for sampling from the real-world data set. (B) The median loss across simulated campaigns with the C-N cross-coupling data base on four different pessimistic buffer lengths and serial sampling as a function of the number of experiments. Each plot is the result of 80 replicates. The loss is calculated relative to an assumed maximum yield of 100%.

Like the prior optimization campaigns, the pessimistic asynchronous sampling policy was benchmarked relative to a standard serial optimization strategy. In this specific campaign, we leveraged modeling information from the original study, and a random forest regressor was used as the belief model instead of the Gaussian process regressor applied in all prior simulations. Uncertainty was estimated through the standard deviation of all forest member predictions.

Of the approximately 4,000 samples in the database, two sets of conditions result in a yield of 100%. While the parameter space is technically 120 dimensions, it is highly constrained through discretization of the parameters and a limited number of possible chemical feature combinations. Regardless, pessimistic asynchronous sampling demonstrated a significant improvement over equivalent serial sampling on this system. Shown in Figure 5B, the pessimistic policy outperformed the serial policy by experiment number for all tested buffer lengths below nine. The highest performing policy, pessimistic sampling with one buffer, reached the optimal yield in 60% fewer samples than serial sampling. While a one sample buffer improved the sampling efficiency over all other methods, increasing the buffer size decreased the efficiency of the policies. The nine-sample buffer performed equivalently to serial sampling with respect to experiment number. Evaluating these results with consideration of the faster sampling rate of asynchronous policies would increase this advantage further.

This result not only indicates that asynchronous pessimistic policies can effectively navigate real-world systems, but it also shows viability in discrete numerical and other constrained spaces. Additionally, the observed improvement over serial methods while using a random forest belief model could indicate further that asynchronous pessimism helps to alleviate deficiencies in uncertainty estimation. The use of ensemble member variance as an uncertainty estimator in a non-parametric ensemble would likely not provide an optimal estimator of uncertainty. Despite this, the asynchronous policy performed favorably in a high complexity surrogate space. Further exploration and development of these methods could reduce the needs for accurate uncertainty estimates and enable the effective application of different models.

## Conclusions

The asynchronous sampling policies presented in this work provide valuable advancement over existing Bayesian optimization strategies for high-cost experimentation in serial experimental systems for both simulated and real-world spaces. While the pessimistic asynchronous sampling policy suffered a performance penalty in low dimension spaces, it performed considerably better than serial or realistic asynchronous policies in parameter spaces with dimensionalities greater than four. Additionally, the performance gap between serial and pessimistic asynchronous policies appears to increase as the dimensionality of the parameter space increases. Furthermore, we have demonstrated that this approach is robust in both systems with sampling noise and real-world collected data sets. The algorithm presented in this work is pertinent to experimental spaces that are relevant to autonomous experimentation, i.e. high-dimensional spaces with noise.

Disregarding the increased sampling rate of asynchronous policies, pessimistic policies may offer greater performance for Bayesian optimization algorithms in high-cost sampling systems. When considering the increased sampling rates, pessimistic policies provide a considerable advantage over existing realistic asynchronous and serial sampling approaches. Further implementation and development of the methods presented here could result in more efficient algorithm driven experimentation and more effective parallelization of experimental processes.

# Supporting Information

## Pessimistic asynchronous sampling in high-cost Bayesian optimization

Amanda A. Volk[1]; Kristofer G. Reyes[2], Jeffrey G. Ethier[3], Luke A. Baldwin[3]

*[1] National Research Council Research Associate, Air Force Research Laboratory, Wright-Patterson Air Force Base, OH, 45433, United States*

*[2] Department of Materials Design and Innovation, University at Buffalo, Buffalo, NY 14260*

*[3] Materials and Manufacturing Directorate, Air Force Research Laboratory, Wright-Patterson Air Force Base, OH 45433, United States*

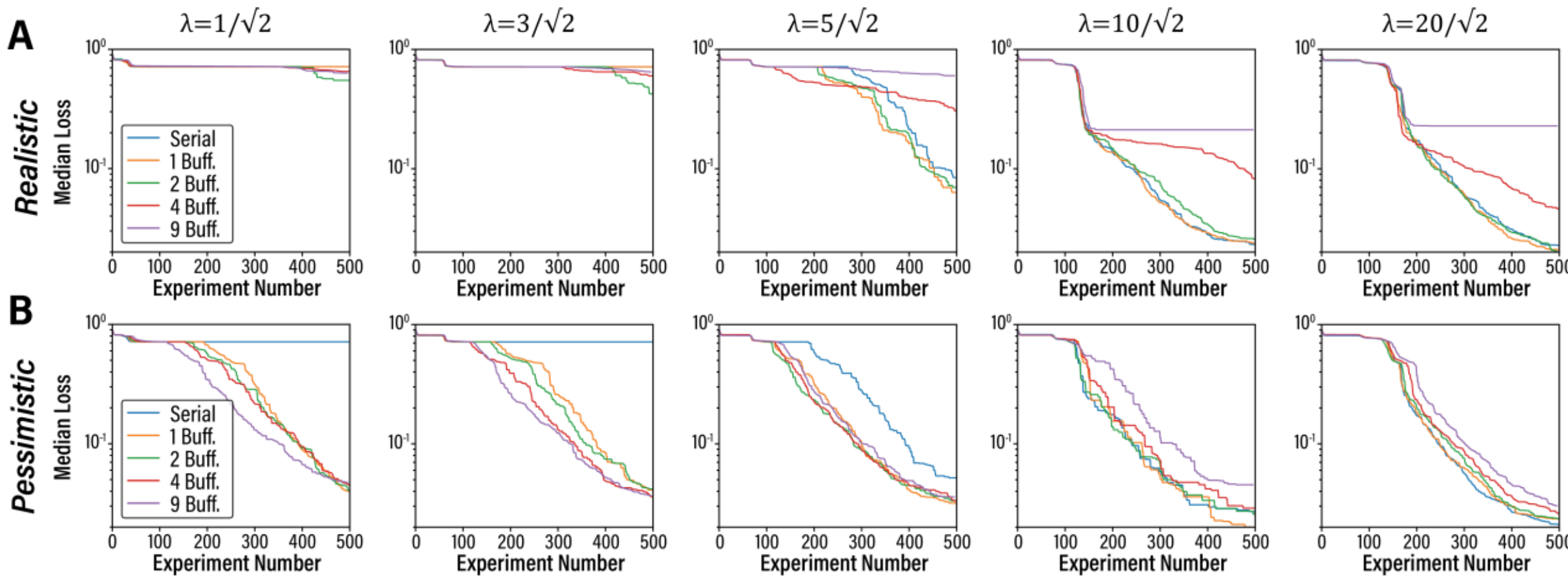

**Figure S.1.** Simulation results of pessimistic and realistic asynchronous decision policies on five-dimensional *TriPeak*. The median loss across all 200 randomized simulated campaigns as a function of the number of experiments for the (A) realistic and (B) pessimistic prediction policies, with (columns) increasing exploration constants, lambda values, in the upper confidence bounds decision policy. The no buffer replicates were repeated for each of the policies and lambda values.

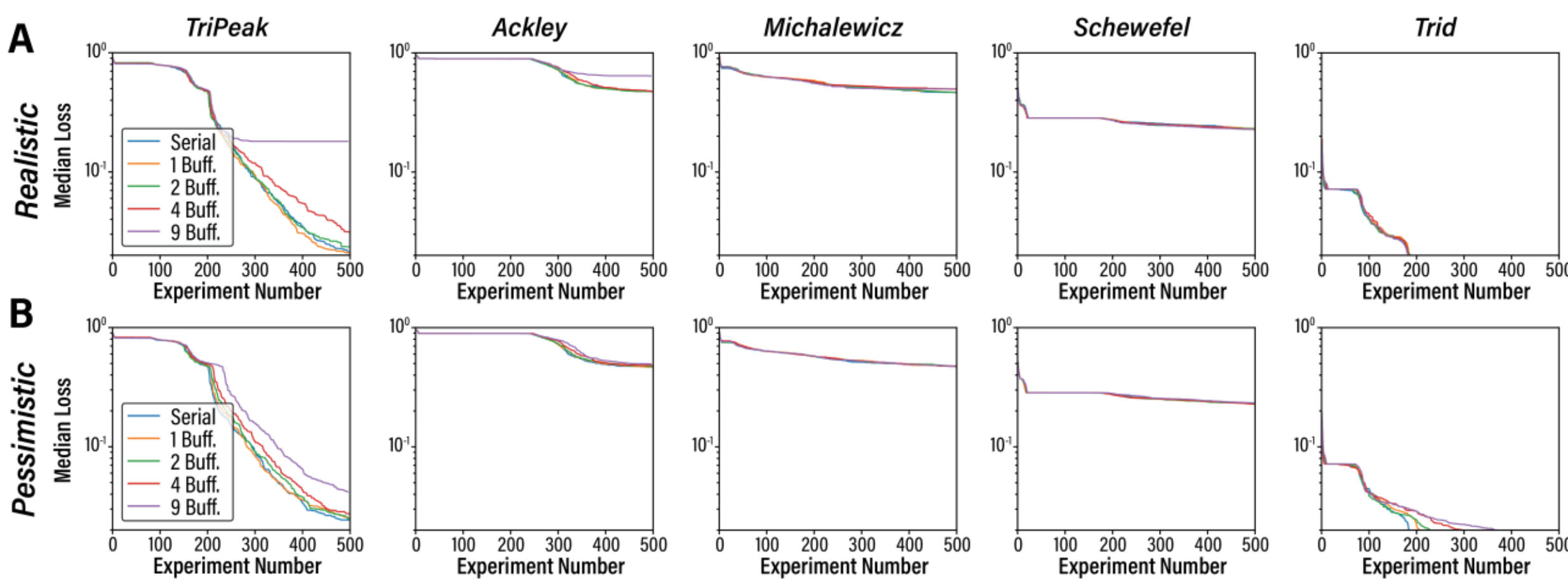

**Figure S.2.** Logarithmic lambda simulation results of pessimistic and realistic asynchronous decision policies on five-dimensional *TriPeak*, *Ackley*, *Michalewicz*, *Schewefel*, and *Trid*. The median loss across all 200 randomized simulated campaigns as a function of the number of experiments for the (A) realistic and (B) pessimistic prediction policies, with a logarithmic upper confidence bounds decision policy. The exploration constant, lambda, is defined as $\lambda = 2\ln\left(\frac{Dt^2\pi^2}{3d}\right)$, where $D$ is the dimensionality of the input space, $t$ is the current experiment number, and $d$ is equal to 1. The no buffer replicates were repeated for each of the policies.

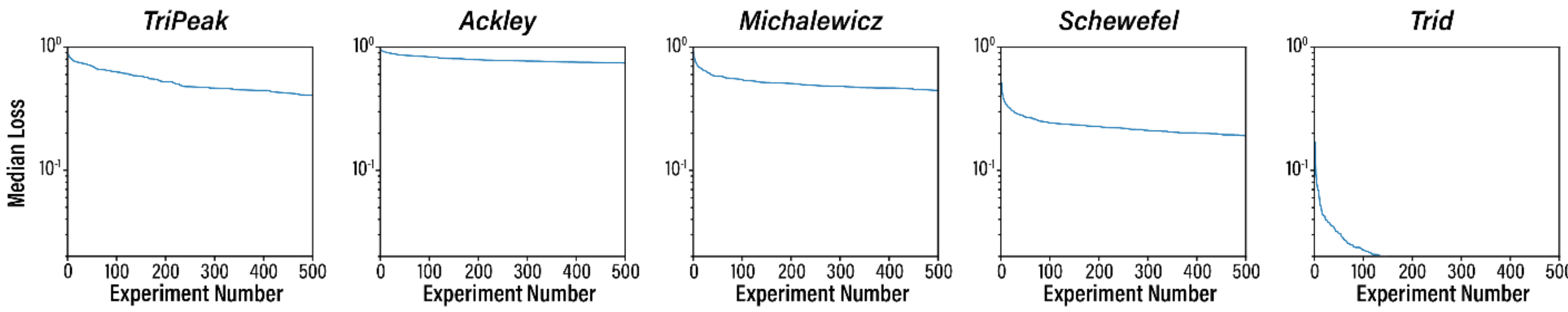

**Figure S.3.** Random sampling control group on five-dimensional *TriPeak*, *Ackley*, *Michalewicz*, *Schewefel*, and *Trid*. The median loss across all 200 randomized simulated campaigns as a function of the number of experiments.

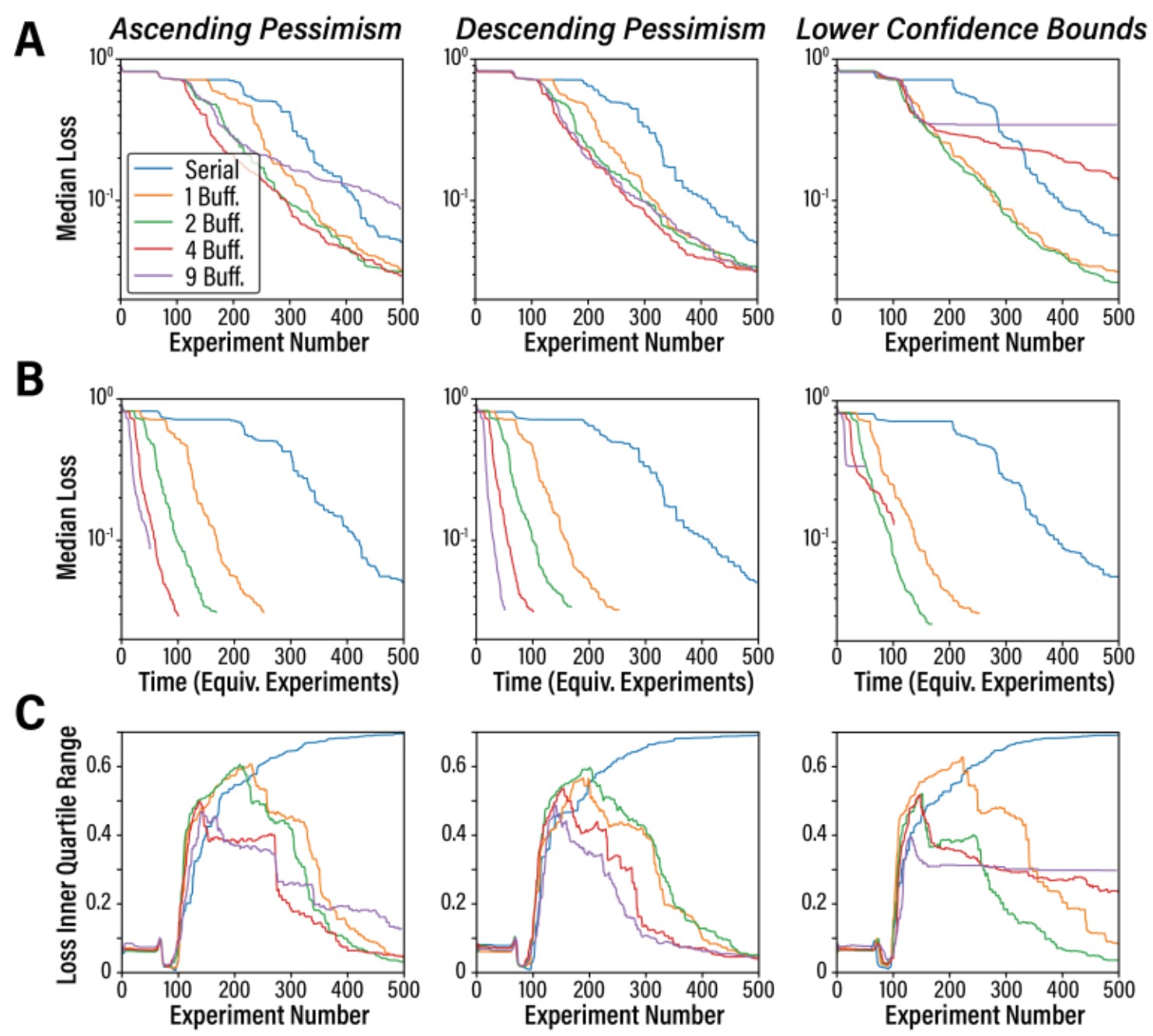

**Figure S.4.** Simulation results of three asynchronous decision policies utilizing different prediction strategies on five-dimensional *TriPeak*. The median loss across all 200 randomized simulated campaigns as a function of (A) the number of experiments and (B) the effective optimization time relative to a single experiment and (C) the inner quartile range of the loss as a function of experiment number across the four decision policies, (first column) ascending pessimism, (second column) descending pessimism, and (third column) lower confidence bound based prediction. Serial replicates were repeated for each of the four policies.

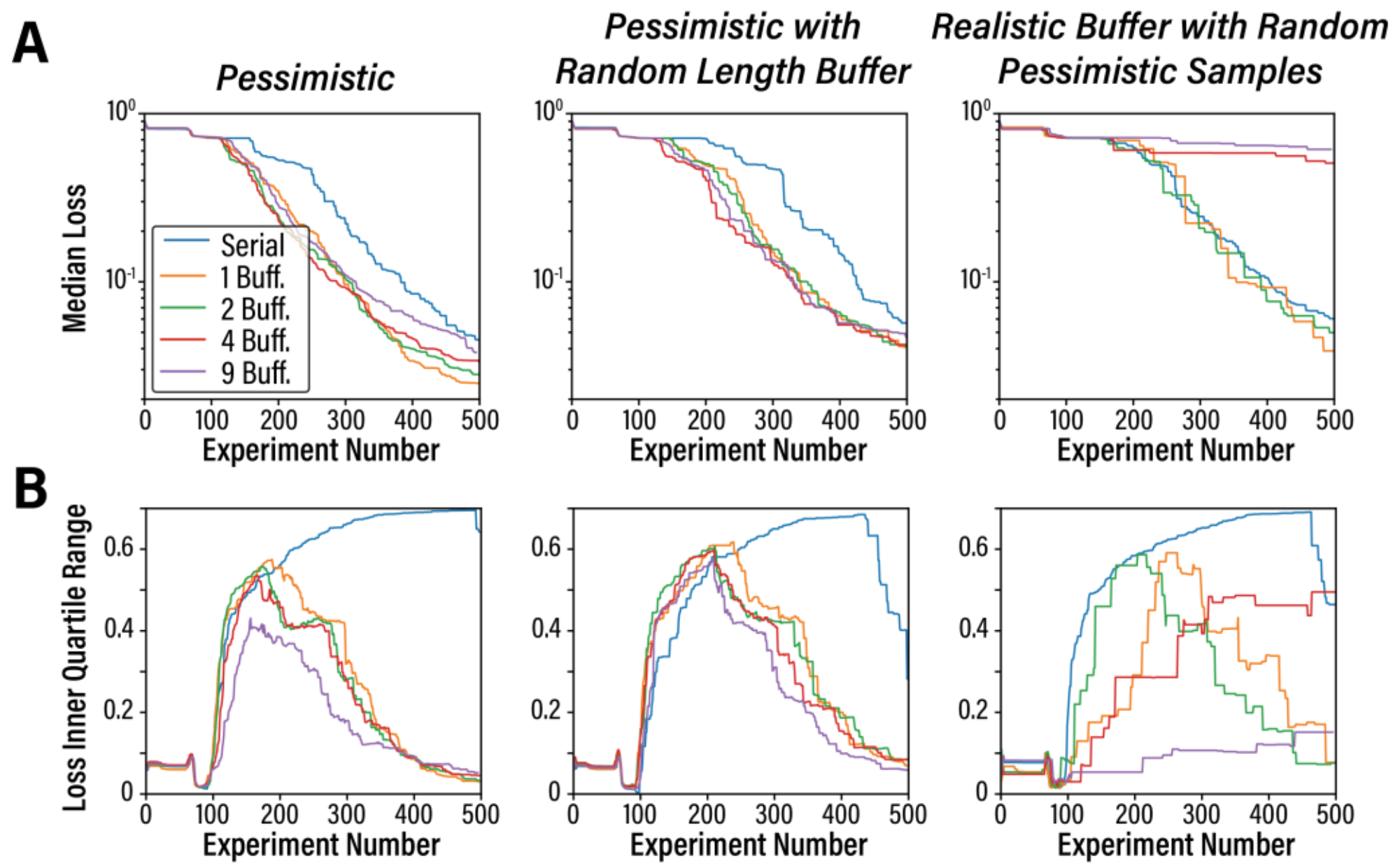

**Figure S.5.** Simulation results of three asynchronous decision policies utilizing different pessimism-based prediction strategies on five-dimensional *TriPeak*. The median loss across all 140 to 200 randomized simulated campaigns as a function of (A) the number of experiments and (B) the inner quartile range of the loss as a function of experiment number across the four decision policies, (first column) pessimistic, (second column) pessimistic with a random length buffer, and (third column) realistic buffer with randomly sampled pessimism. The random length buffer was selected uniformly between 0 and the specified buffer length. The pessimistic samples were formed by randomly assigning a pessimistic value to a number of samples equal to the current buffer length. Serial replicates were repeated for each of the four policies.